\documentclass[fullpaper,final]{nldl}

\paperID{57}

\vol{352}

\usepackage{mathtools}

\usepackage{graphicx}
\usepackage{fontawesome5}

\usepackage{booktabs}

\usepackage{enumitem}

\usepackage{algorithm}
\usepackage{algorithmic}

\usepackage{xcolor}

\usepackage{listings}
\usepackage{hyperref}
\usepackage{url}
\hypersetup{
  pdfusetitle,
  colorlinks,
  linkcolor = BrickRed,
  citecolor = NavyBlue,
  urlcolor  = Magenta!80!black,
}

\usepackage{amsmath,amssymb,amsthm}

\newcommand{\norm}[1]{\left\lVert #1 \right\rVert}

\title{
EHR Representation Learning with Multi-Granularity\\ Cauchy-Schwarz Contrastive Learning
}
\author[1]{Keyne Kassapa Oei}
\author[1,2]{Shujian Yu}
\author[1]{Qing Liu}
\author[1,3]{Robert Jenssen}
\affil[1]{Department of Physics and Technology, UiT The Arctic University of Norway}
\affil[2]{Department of Computer Science, Vrije Universiteit Amsterdam}
\affil[3]{Department of Computer Science, University of Copenhagen}
\affil[1]{\texttt{keyne.k.oei@uit.no}}

\begin{document}
\maketitle

\begin{abstract}
Self-supervised EHR models commonly reconstruct masked medical codes or predict future events, without explicitly modeling clinical relations across patients and visits.
Contrastive methods can introduce relational supervision but usually reduce these relations to binary positive and negative pairs.
We introduce EHR-CSS, a multi-granularity self-supervised contrastive framework for patient, visit, and medical-code representations.
The patient and visit objectives assign target weights according to clinical similarity, while the medical-code objectives assign target weights according to the medical codes to be reconstructed or predicted.
At each granularity, the proposed Cauchy--Schwarz Sequence (CSS) loss aligns a model distribution over candidate representations with its target distribution and includes a uniformity term over the candidate representations.
Across evaluations on EHRSHOT with few-shot labels and on MIMIC-IV with fully supervised fine-tuning, EHR-CSS achieves the best overall performance among the evaluated methods.
The code is available at \url{https://github.com/keynekassapa13/EHR-CSS}.
\end{abstract}

\section{Introduction}
\label{sec_introduction}

\begin{figure}[!t]
\centering
\includegraphics[width=\linewidth]{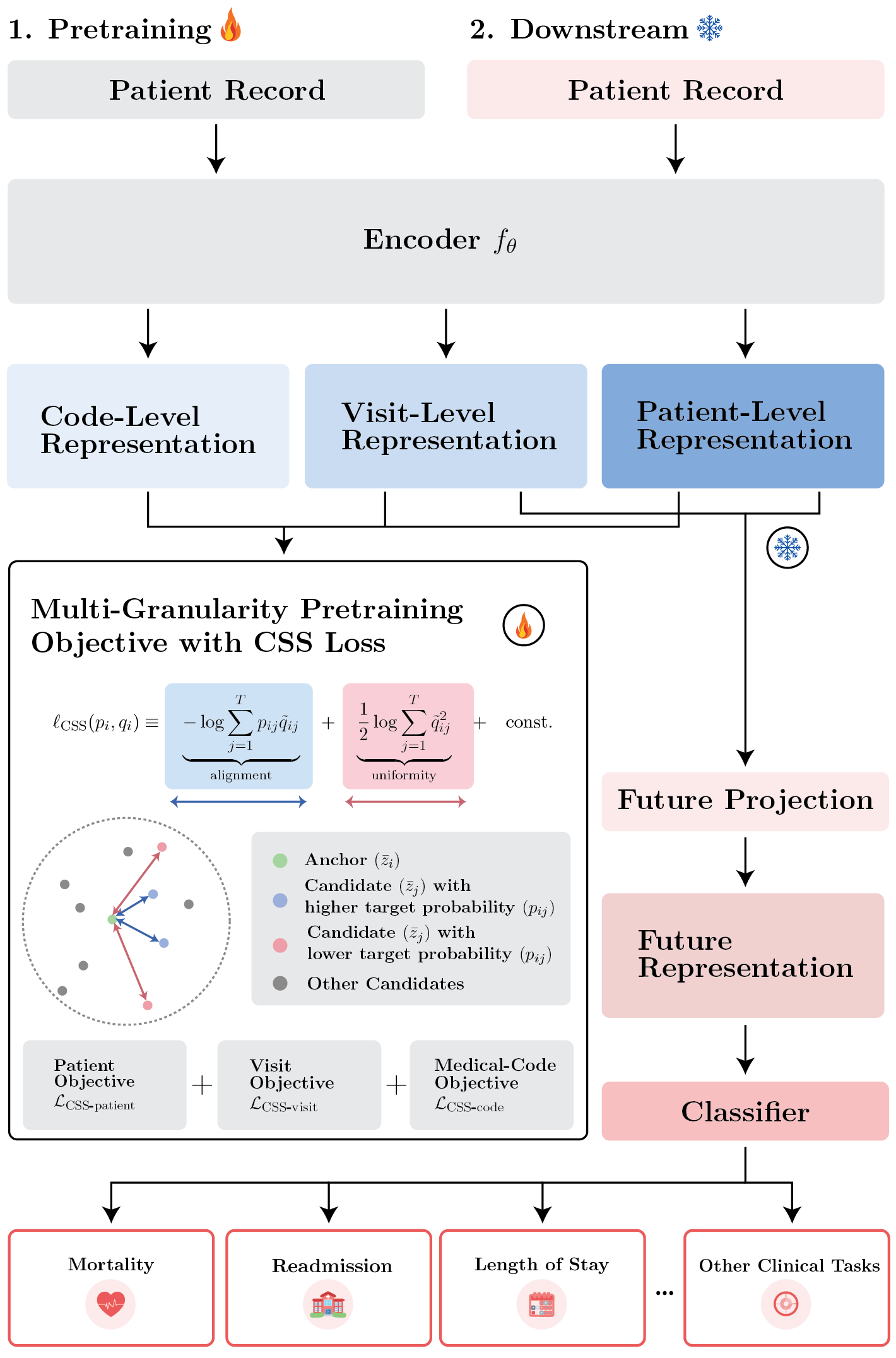}
\caption{
Overview of EHR-CSS pretraining and downstream evaluation.
The CSS objective trains medical-code, visit, and patient representations.
\textcolor{orange}{\faFire} denotes parameters optimized during pretraining.
\textcolor{blue}{\faSnowflake} denotes components frozen during EHRSHOT \cite{ehrshot} downstream evaluation.
}
\label{fig:ehr-css-overview}
\end{figure}

Predictive modeling with longitudinal electronic health records (EHRs) has become increasingly important for clinical risk assessment and outcome prediction from patient histories \cite{zheng2025ehrreview,hama2025ehrprediction}.
State-of-the-art methods for self-supervised EHR representation learning broadly follow generative and contrastive training strategies \cite{zheng2025ehrreview}.
Generative methods reconstruct masked medical events \cite{behrt,medbert,cehrbert} or predict future events \cite{clmbr,ehrshot,transformehr,motor,ethos,ehrmamba}, whereas contrastive methods learn by comparing representations of related and unrelated examples \cite{zheng2025ehrreview,wang2020alignment}.
Despite this progress, existing EHR representation learning methods leave two challenges unresolved.
First, masked-event \cite{behrt,medbert,cehrbert} and future-event \cite{clmbr,ehrshot,transformehr,motor,ethos,ehrmamba} objectives derive self-supervised targets only from events within each patient record and therefore do not explicitly model clinical relations across patients or visits.
Second, relational approaches based on patient-similarity objectives, graphs, or hypergraphs \cite{suo2018similarity,hcl,xu2023hypehr,gatorclr} often depend on outcome labels or specialized graph structures.
These challenges motivate our hypothesis that clinical relations at the patient and visit granularities can complement prediction-based self-supervision at the medical-code granularity.
Contrastive learning is suitable for this setting because it can use relations among examples as training signals without requiring outcome labels or explicit graph structures.

Conventional contrastive objectives represent such relations as positive and negative pairs between an anchor representation and candidate representations \cite{wang2020alignment,wanyan2021positive,softclt}.
Existing EHR contrastive methods define positive pairs using outcome labels \cite{scehr}, feature-space neighborhoods or predefined patient similarity measures \cite{wanyan2021positive}, or augmented views of the same patient record \cite{king2024ehrcontrast,tact}.
However, binary pair assignments cannot represent different degrees of similarity and typically treat candidates outside the positive set as negatives.

To represent these graded clinical relations directly, we propose EHR-CSS, a multi-granularity contrastive framework for structured longitudinal EHRs.
Figure~\ref{fig:ehr-css-overview} summarizes EHR-CSS pretraining and downstream evaluation.
The model sums the medical-code, age, segment, and visit-position embeddings before passing the sequence to the shared encoder.
The encoder outputs form the medical-code representations and are averaged within and across visits to produce visit and patient representations.
Each objective compares an anchor representation with candidate representations at the same granularity.
The model distribution assigns probability to each candidate according to its learned similarity to the anchor representation.
The target distribution assigns probability to the same candidate according to the self-supervised signal defined for each objective.
The CSS loss adapts the established CS divergence to align each model distribution with its target distribution.
We choose CS divergence because KL divergence uses pointwise probability ratios that can become unstable near zero \cite{principe2000itl,jenssen2006cs,zhang2025crossmodal}.
CSS combines an alignment term that favors candidates supported by the target distribution with a uniformity term over the candidate representations.
Each training example contains an observed patient history, with the following visit used as a prediction target.
The patient and visit objectives derive target weights from medical-code co-occurrence, while the three medical-code objectives target a masked medical code, the medical codes in the following visit, and the next medical code in the sequence.

We evaluate EHR-CSS across few-shot and fully supervised downstream settings.
EHRSHOT~\cite{ehrshot} evaluates the frozen representation across diverse few-shot tasks, while MIMIC-IV~\cite{mimiciv} evaluates fully supervised fine-tuning for clinical outcomes.
Across these evaluations, EHR-CSS achieves the best overall performance among the state-of-the-art methods.

Our contributions are summarized as follows.
\begin{itemize}
    \item We construct EHR-specific patient and visit target distributions that encode graded clinical similarity from medical-code identity and co-occurrence, with elapsed time additionally informing the visit target distributions.

    \item We introduce EHR-CSS, a multi-granularity pretraining framework that trains the patient and visit objectives defined by these target distributions jointly with masked-code, next-visit code, and dense next-code prediction. The Cauchy--Schwarz Sequence (CSS) loss adapts the established CS divergence \cite{principe2000itl,jenssen2006cs} to these five model and target distribution pairs rather than introducing a new divergence.

    \item We show that EHR-CSS achieves the best overall performance on the EHRSHOT~\cite{ehrshot} and MIMIC-IV~\cite{mimiciv} benchmarks.
\end{itemize}

\section{Related Work}
\label{sec:related-work}
\begingroup
\setlength{\parindent}{0pt}
\setlength{\parskip}{0.6\baselineskip}

\textbf{Masked-event prediction.}
Masked-event pretraining learns EHR representations by reconstructing medical codes removed from the observed patient history.
BEHRT \cite{behrt} applies a bidirectional Transformer to diagnosis sequences represented by medical-code, age, position, and visit-segment embeddings.
Med-BERT \cite{medbert} extends masked-code pretraining to large diagnosis vocabularies and adds prolonged hospital stay prediction.
CEHR-BERT \cite{cehrbert} represents elapsed time through time-interval tokens and temporal embeddings and adds visit-type prediction to masked-code prediction.

\textbf{Future-event prediction.}
Future-event pretraining instead learns from events that follow the observed patient history.
CLMBR \cite{clmbr} uses a recurrent encoder to predict the next-day medical-code set, whereas CLMBR-T \cite{ehrshot} uses a causal Transformer to predict the next medical code.
TransformEHR \cite{transformehr} uses an encoder-decoder Transformer to generate the complete diagnosis-code sequence of the next visit.
MOTOR \cite{motor} uses time-to-event pretraining to estimate when selected clinical events will occur.
ETHOS \cite{ethos} serializes heterogeneous clinical events, numerical values, and time intervals into a common patient timeline and trains a causal Transformer to generate its continuation.
EHRMamba \cite{ehrmamba} combines next-token pretraining with Mamba state-space blocks and prompted fine-tuning across multiple prediction tasks.

\textbf{Contrastive and relational learning.}
Contrastive EHR methods derive relations from outcome labels, patient-similarity measures, or augmented patient histories.
SCEHR \cite{scehr} uses outcome labels to relate patient representations, while BYOS-EHR \cite{wanyan2021positive} selects positive candidates through learned feature similarity or a predefined patient-similarity measure.
\citet{king2024ehrcontrast} combine masked-value regression with sequence-level contrastive learning, whereas TACT \cite{tact} constructs augmented patient histories using diagnosis and medication hierarchies.
Related EHR methods encode clinical relations through supervised patient-similarity objectives \cite{suo2018similarity} or graph and hypergraph structures \cite{hcl,xu2023hypehr,gatorclr}.
Multi-granularity methods learn medical-code, visit, and patient representations by predicting co-occurring medical codes \cite{med2vec} or outcome labels \cite{mlrl}.
EHR-CSS instead trains patient and visit representations directly with clinical-similarity target distributions and requires no outcome labels.

\textbf{Representation learning beyond EHRs.}
Prior work on sequential representation learning motivates replacing binary pair assignments with continuous similarity weights.
Video representation learning uses temporal weighting \cite{carl} or differentiable local sequence alignment \cite{oei2024lac}.
SoftCLT derives continuous weights from sequence distance and timestamp differences \cite{softclt}.
EHR-CSS instead derives graded weights from medical-code identity, medical-code co-occurrence, and irregular visit timing.
A separate line of work uses CS divergence for representation learning and distribution alignment \cite{jenssen2006cs,tran2022csrae,yu2023conditional,yin2024domain}.
\citet{zhang2025crossmodal} apply CS divergence to align feature distributions for cross-modal retrieval.
EHR-CSS uses the same CS divergence but defines separate model and target distributions over candidate patients, visits, and medical codes.
\par\endgroup
 
\begin{figure*}[t]
\centering
\includegraphics[width=\textwidth]{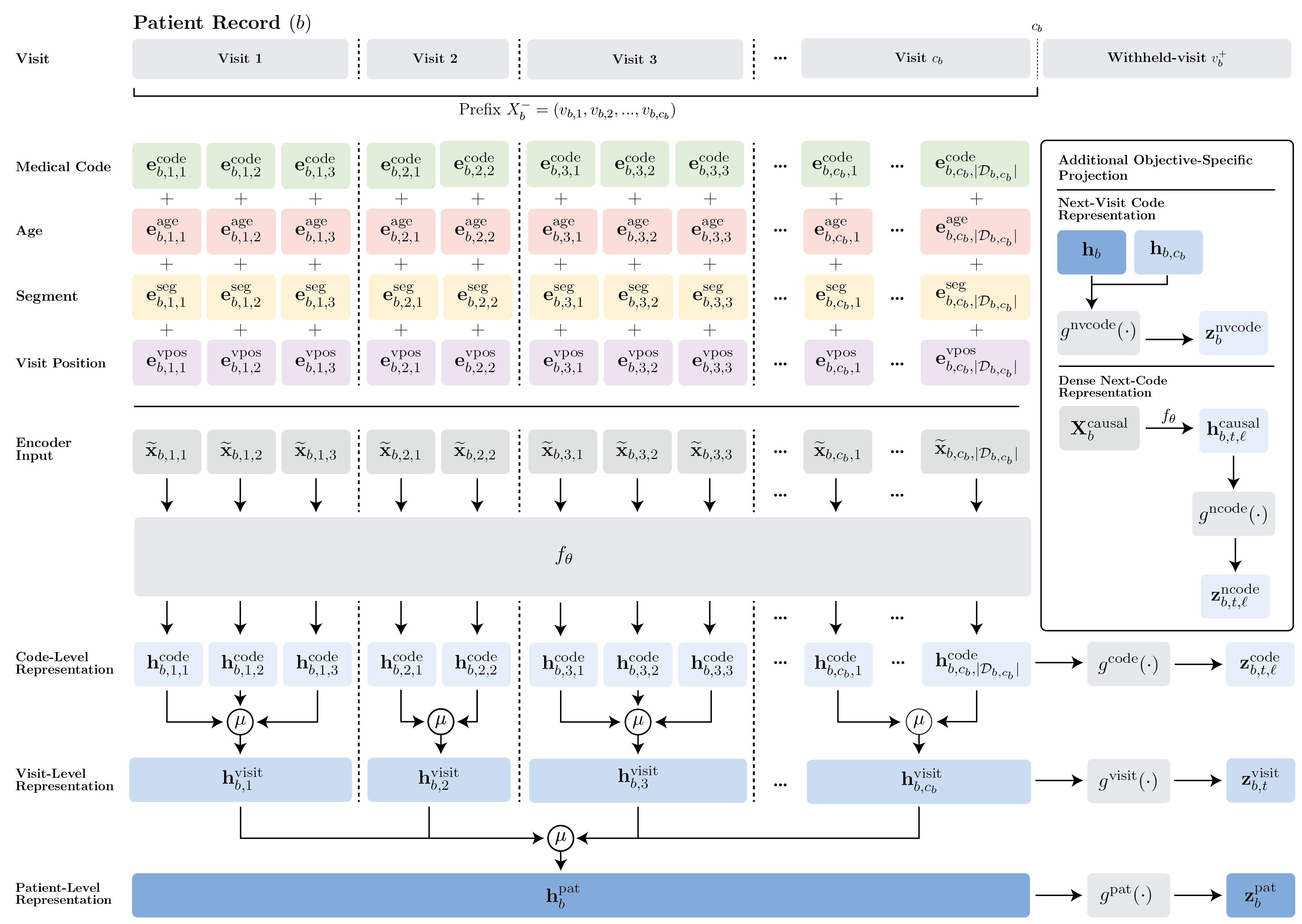}
\caption{
EHR-CSS input and multi-granularity representation pipeline. 
The symbol $\mu$ denotes average.
The observed prefix is encoded into medical-code, visit, and patient representations.
Projection networks map these representations to the five objective-specific anchor spaces.
The withheld visit defines the next-visit target.
A causal pass through the shared encoder supports dense next-code prediction.
}
\label{fig:ehr-css-pipeline}
\end{figure*}

\section{Method}
\label{sec:method}

Figure~\ref{fig:ehr-css-pipeline} illustrates the EHR-CSS pretraining framework, including the input and representation construction detailed in Section~\ref{sec:ehr-representation}.
Section~\ref{sec:general-objective} defines the shared CSS objective over an anchor representation and its candidate representations.
Section~\ref{sec:training-objectives} specifies the five model and target distribution pairs, while Section~\ref{sec:downstream-representation} defines the future representation used for downstream prediction.

\subsection{Multi-Granularity EHR Representations}
\label{sec:ehr-representation}

\subsubsection{Observed-prefix construction}

For patient $b$, let $X_b=(v_{b,1},\ldots,v_{b,m_b})$ denote the ordered sequence of $m_b$ visits.
Each visit is represented as
\begin{equation}
v_{b,t}
=
\left(\mathcal D_{b,t} \, , \, \text{age}_{b,t}  \, , \, t_{b,t}^{\mathrm{day}}\right),
\end{equation}
where $\mathcal D_{b,t}$ is the set of medical codes recorded at visit $t$ from a vocabulary of $V$ medical codes, $\text{age}_{b,t}$ is the patient's age, and $t_{b,t}^{\mathrm{day}}$ is the cumulative day coordinate of the visit.
For each dataset, we fix an interval of fractions of $m_b$ and draw the cutoff $c_b$ uniformly from the visit positions inside this interval.
We resample $c_b$ at each epoch and define the observed prefix as
\begin{equation}
X_b^{-}=(v_{b,1}\, , \, \ldots \, , \, v_{b,c_b}).
\label{eq:prefix-next-visit}
\end{equation}
The following visit is withheld as
\begin{equation}
v_b^{+}=v_{b,c_b+1}.
\label{eq:withheld-next-visit}
\end{equation}
The encoder receives the observed prefix $X_b^{-}$, while the medical codes in $v_b^{+}$ define the next-visit target distribution.

\subsubsection{Sequence encoding}

\paragraph{Medical-code-level representations.} 

Similar to BEHRT \cite{behrt}, we learn look-up tables that map each medical code, age, and visit position in the EHR sequence to an embedding.
For the $\ell$-th medical-code belonging to $\mathcal{D}_{b,t}$, we denote its embedding as $\mathbf{e}^{\mathrm{code}}_{b,t,\ell}$ and embeddings for patient's age and visit-position as $\mathbf{e}^{\mathrm{age}}_{b,t,\ell}$ and $ \mathbf{e}^{\mathrm{vpos}}_{b,t,\ell}$, respectively.
To indicate which visit each medical code belongs to, we introduce a segment embedding, denoted by $\mathbf{e}^{\mathrm{seg}}_{b,t,\ell}$.
All embeddings are initialized from a Gaussian distribution with zero mean and a standard deviation of $0.02$.
We combine the medical-code, age, segment, and visit-position embeddings as
\begin{equation}
    \widetilde{\mathbf{x}}_{b,t,\ell}
    =
    \mathbf{e}^{\mathrm{code}}_{b,t,\ell}
    + \mathbf{e}^{\mathrm{age}}_{b,t,\ell}
    + \mathbf{e}^{\mathrm{seg}}_{b,t,\ell}
    + \mathbf{e}^{\mathrm{vpos}}_{b,t,\ell}\;.
    \label{eq:input-embedding}
\end{equation}
We use a bidirectional Transformer $f_\theta$ as the shared sequence encoder and apply it to the combined embeddings to obtain the medical-code-level representations

{
\footnotesize
\begin{align}
&\left(
\mathbf{h}^{\mathrm{code}}_{b,1,1}, \ldots, \mathbf{h}^{\mathrm{code}}_{b,1,|\mathcal{D}_{b,1}|},
\ldots,
\mathbf{h}^{\mathrm{code}}_{b,c_b,1}, \ldots, \mathbf{h}^{\mathrm{code}}_{b,c_b,|\mathcal{D}_{b,c_b}|}
\right)
\nonumber\\
&=
f_{\theta}\!\left(
\widetilde{\mathbf{x}}_{b,1,1}, \ldots, \widetilde{\mathbf{x}}_{b,1,|\mathcal{D}_{b,1}|},
\ldots,
\widetilde{\mathbf{x}}_{b,c_b,1}, \ldots, \widetilde{\mathbf{x}}_{b,c_b,|\mathcal{D}_{b,c_b}|}
\right)\;,
\end{align}
}

\noindent
where $\mathbf{h}^{\mathrm{code}}_{b,t,\ell}$ is the representation for the $\ell$-th medical code in $\mathcal{D}_{b,t}$ at the $t$-th visit of patient $b$ and $|\cdot|$ is the cardinality of a set.

Alongside the bidirectional representations, we use the same encoder with left-to-right attention to obtain a second medical-code representation at each position.
This representation, denoted by $\mathbf{h}^{\mathrm{causal}}_{b,t,\ell}$, uses only the current and preceding medical-code positions to predict the following medical code.

\paragraph{Visit- and patient-level representations.} We obtain the visit-level representation $\mathbf{h}^{\mathrm{visit}}_{b, t}$ via averaging medical-code-level representations belonging to the same visit and obtain the patient-level representation $\mathbf{h}^{\mathrm{pat}}_b$ via averaging visit representations belonging to the same patient:
\begin{equation}
    \mathbf{h}^{\mathrm{visit}}_{b, t} = \frac{1}{|\mathcal{D}_{b, t}|} \sum_{\ell=1}^{|\mathcal{D}_{b,t}|}\mathbf{h}^{\mathrm{code}}_{b,t,\ell}\;, \quad \mathbf{h}^{\mathrm{pat}}_b=\frac{1}{c_b}\sum_{t=1}^{c_b} \mathbf{h}^{\mathrm{visit}}_{b, t}\;.
\label{eq:hierarchical-pooling}
\end{equation}

\subsubsection{Objective-specific projection}
Each objective uses a projection network to map its input representation to its objective-specific space.
The patient and visit projection networks use two linear layers separated by a rectified linear unit (ReLU) activation \cite{nair2010rectified}.
The masked-code, next-visit, and dense next-code projection networks use a linear layer followed by a Gaussian error linear unit (GELU) activation \cite{hendrycks2016gelu} and layer normalization \cite{ba2016layernorm}.
The projected anchor representations for the patient, visit, masked-code, next-visit code, and dense next-code objectives are defined as 
\begin{equation}
\begin{aligned}
\mathbf{z}_b^{\mathrm{pat}}
&=g^{\mathrm{pat}}(\mathbf{h}_b^{\mathrm{pat}}),
\\
\mathbf{z}_{b,t}^{\mathrm{visit}}
&=g^{\mathrm{visit}}(\mathbf{h}_{b,t}^{\mathrm{visit}}),
\\
\mathbf{z}_{b,t,\ell}^{\mathrm{code}}
&=g^{\mathrm{code}}(\mathbf{h}_{b,t,\ell}^{\mathrm{code}}),
\\
\mathbf{z}_b^{\mathrm{nvcode}}
&=g^{\mathrm{nvcode}}(
[\mathbf{h}_b^{\mathrm{pat}};\mathbf{h}_{b,c_b}^{\mathrm{visit}}]
),
\\
\mathbf{z}_{b,t,\ell}^{\mathrm{ncode}}
&=g^{\mathrm{ncode}}(\mathbf{h}_{b,t,\ell}^{\mathrm{causal}}).
\end{aligned}
\label{eq:objective-projections}
\end{equation}
Here, $[\cdot;\cdot]$ denotes concatenation.
Every projected representation is normalized to unit length, denoted by $\bar{\mathbf{z}}^s$, where $s$ indexes the corresponding objective.

\subsection{Cauchy--Schwarz Sequence (CSS) Loss}
\label{sec:general-objective}

We define the CSS loss for an anchor representation and its candidate representations.
The index $i$ identifies the anchor representation, $j=1,\ldots,T$ indexes its candidate representations, and $w_{ij}\geq 0$ is the target weight assigned to candidate $j$ for anchor $i$.
We define the target distribution $\mathbf{p}_i=(p_{i1},\ldots,p_{iT})$ by normalizing these weights as
\begin{equation}
p_{ij}
=
\frac{w_{ij}}{\sum_{k=1}^{T}w_{ik}},
\qquad j=1,\ldots,T.
\label{eq:target-normalization}
\end{equation}
The target weights are constructed from observed clinical relations or prediction targets and remain fixed with respect to the model parameters.

Let \(\bar{\mathbf{z}}_i\) and \(\bar{\mathbf{z}}_j\) denote the unit-normalized anchor representation and candidate representation, respectively.
Their squared distance defines the model distribution \(\mathbf{q}_i=(q_{i1},\ldots,q_{iT})\) as
\begin{equation}
q_{ij}
=
\frac{
\exp\!\left(
-\frac{1}{\tau}\norm{\bar{\mathbf{z}}_i-\bar{\mathbf{z}}_j}^{2}
\right)
}{Z_i},
\label{eq:model-pmf}
\end{equation}
where \(\tau>0\) is the temperature and
\begin{equation}
Z_i
=
\sum_{k=1}^{T}
\exp\!\left(
-\frac{1}{\tau}\norm{\bar{\mathbf{z}}_i-\bar{\mathbf{z}}_k}^{2}
\right)
\label{eq:partition}
\end{equation}
normalizes the model weights across the candidate representations.

We define the CSS loss via the CS divergence \cite{principe2000itl,jenssen2006cs} between the target distribution and the model distribution as follows
\begin{equation}
\ell_{\mathrm{CSS}}(\mathbf{p}_i,\mathbf{q}_i)
=
-\log
\frac{
\sum_{j=1}^{T}p_{ij}q_{ij}
}{
\left(
\sum_{j=1}^{T}p_{ij}^2
\right)^{1/2}
\left(
\sum_{j=1}^{T}q_{ij}^2
\right)^{1/2}
}.
\label{eq:cs-objective}
\end{equation}
The CS divergence is invariant to positive rescaling of either argument \cite{jenssen2006cs,jenssen2024mapit}.
Consequently, the partition function \(Z_i\) cancels from Eq.~\eqref{eq:cs-objective}.
Define the unnormalized model-weight vector as $\tilde{\mathbf{q}}_i=(\tilde q_{i1},\ldots,\tilde q_{iT})$, where
\begin{equation}
\tilde q_{ij}
=
\exp\!\left(
-\frac{1}{\tau}\norm{\bar{\mathbf{z}}_i-\bar{\mathbf{z}}_j}^{2}
\right),
\qquad
q_{ij}=\frac{\tilde q_{ij}}{Z_i}.
\label{eq:unnormalized-model}
\end{equation}
The CSS loss can then be evaluated without the partition function
\begin{equation}
\ell_{\mathrm{CSS}}(\mathbf{p}_i,\mathbf{q}_i)
= -\log
\frac{
\sum_{j=1}^{T}p_{ij}\tilde q_{ij}
}{
\left(
\sum_{j=1}^{T}p_{ij}^2
\right)^{1/2}
\left(
\sum_{j=1}^{T}\tilde q_{ij}^2
\right)^{1/2}
}.
\label{eq:cs-unnormalized}
\end{equation}

The CSS loss decomposes into alignment and uniformity terms
\begin{equation}
\ell_{\mathrm{CSS}}(\mathbf{p}_i,\mathbf{q}_i)
=
\underbrace{
-\log\sum_{j=1}^{T}p_{ij}\tilde q_{ij}
}_{\text{alignment}}
+
\underbrace{
\frac{1}{2}\log\sum_{j=1}^{T}\tilde q_{ij}^{2}
}_{\text{uniformity}}
+C_i,
\label{eq:alignment-uniformity}
\end{equation}
where \(C_i=\frac{1}{2}\log\sum_{j=1}^{T}p_{ij}^{2}\) is constant with respect to the model parameters.
Appendix~\ref{app:css-decomposition} provides the complete derivation of Eq.~\eqref{eq:alignment-uniformity}.
The alignment term increases the unnormalized model weight assigned to candidate representations favored by the target distribution.
The uniformity term penalizes a large squared norm of the model weights and discourages high similarity between the anchor representation and many candidate representations.

\subsection{Multi-Granularity Pretraining Objective}
\label{sec:training-objectives}

\subsubsection{Full Pretraining Objective}
\label{sec:model-distributions}
\label{sec:targets}

We group the five losses according to the granularity of their candidate representations.
The coefficients $\lambda$ weight the individual objectives.
The complete multi-granularity objective is
\begin{equation}
\mathcal L_{\mathrm{CSS}}
=
\mathcal L_{\mathrm{CSS\text{-}patient}}
+\mathcal L_{\mathrm{CSS\text{-}visit}}
+\mathcal L_{\mathrm{CSS\text{-}code}}.
\label{eq:full-objective}
\end{equation}
Let $\mathcal A_s$ contain the valid anchors for objective $s$.
The patient and visit components are
\begin{equation}
\begin{aligned}
\mathcal L_{\mathrm{CSS\text{-}patient}}
&=
\frac{\lambda_{\mathrm{pat}}}{|\mathcal A_{\mathrm{pat}}|}
\sum_{b\in\mathcal A_{\mathrm{pat}}}
\ell_{\mathrm{CSS}}
\!\left(\mathbf{p}_b^{\mathrm{pat}},\mathbf{q}_b^{\mathrm{pat}}\right),
\\
\mathcal L_{\mathrm{CSS\text{-}visit}}
&=
\frac{\lambda_{\mathrm{vis}}}{|\mathcal A_{\mathrm{vis}}|}
\sum_{(b,t)\in\mathcal A_{\mathrm{visit}}}
\ell_{\mathrm{CSS}}
\!\left(\mathbf{p}_{b, t}^{\mathrm{visit}},\mathbf{q}_{b, t}^{\mathrm{visit}}\right).
\end{aligned}
\label{eq:patient-visit-objectives}
\end{equation}
The three objectives defined over medical-code candidate representations form the code component
{\medmuskip=1mu\thickmuskip=2mu\thinmuskip=1.5mu
\begin{equation}
\begin{aligned}
\mathcal L_{\mathrm{CSS\text{-}code}}
& ={}
\frac{\lambda_{\mathrm{code}}}{|\mathcal A_{\mathrm{code}}|}
\sum_{(b,\ell)\in\mathcal A_{\mathrm{code}}}
\ell_{\mathrm{CSS}}
\!\left(\mathbf{p}_{b, \ell}^{\mathrm{code}},\mathbf{q}_{b, \ell}^{\mathrm{code}}\right)
\\
&+
\frac{\lambda_{\mathrm{nvcode}}}{|\mathcal A_{\mathrm{nvcode}}|}
\sum_{b\in\mathcal A_{\mathrm{nvcode}}}
\ell_{\mathrm{CSS}}
\!\left(\mathbf{p}_b^{\mathrm{nvcode}},\mathbf{q}_b^{\mathrm{nvcode}}\right)
\\
&+
\frac{\lambda_{\mathrm{ncode}}}{|\mathcal A_{\mathrm{ncode}}|}
\sum_{(b,\ell)\in\mathcal A_{\mathrm{ncode}}}
\ell_{\mathrm{CSS}}
\!\left(\mathbf{p}_{b, \ell}^{\mathrm{ncode}},\mathbf{q}_{b, \ell}^{\mathrm{ncode}}\right).
\end{aligned}
\label{eq:code-objective}
\end{equation}
}

\subsubsection{Patient and Visit Objectives}

For anchor patient $b$, candidate patient $i$ is drawn from the current minibatch or patient queue, with $b \neq i$.
For anchor visit $t$, candidate visit $u$ is another nonempty visit in the observed history of patient $b$, with $t \neq u$.
The unnormalized patient and visit model scores are
\begin{equation}
\begin{aligned}
\tilde q_{b,i}^{\mathrm{pat}}
&=
\exp\!\left(
-\frac{1}{\tau}
\norm{
\bar{\mathbf{z}}_b^{\mathrm{pat}}
-\bar{\mathbf{z}}_i^{\mathrm{pat}}
}^{2}
\right),
\\
\tilde q_{b,t,u}^{\mathrm{vis}}
&=
\exp\!\left(
-\frac{1}{\tau}
\norm{
\bar{\mathbf{z}}_{b,t}^{\mathrm{visit}}
-\bar{\mathbf{z}}_{b,u}^{\mathrm{visit}}
}^{2}
\right).
\end{aligned}
\label{eq:patient-visit-model-weights}
\end{equation}

Patients with similar observed histories should have similar patient representations, and clinically similar visits should have similar visit representations.
The patient target distribution $\mathbf{p}_b^{\mathrm{pat}}$ therefore assigns greater probability to patients with clinically similar observed histories.
The visit target distribution $\mathbf{p}_{b,t}^{\mathrm{visit}}$ assigns greater probability to clinically similar visits that are also close in time.
To construct both target distributions, we use clinical similarity derived from medical-code co-occurrence as a supervision signal.

\paragraph{Clinical Similarity Construction.}
\label{sec:clinical-similarity}

We derive the medical-code similarity matrix $\mathbf{A}$ using positive pointwise mutual information (PPMI) estimated from patient-level medical-code co-occurrence \cite{church1990word,bullinaria2007extracting}, following its prior use in EHR representation learning \cite{xiang2019timesensitive,hong2021keser}.
Appendix \ref{app:ppmi} gives the count-based estimator used to obtain $\mathbf{A}$.
We normalize the positive entries in each row to sum to one, yielding $\widetilde{\mathbf{A}}$, and combine this matrix with exact medical-code identity as
\begin{equation}
\mathbf{S}
=
(1-\eta)\mathbf{I}+\eta\widetilde{\mathbf{A}},
\qquad \eta\in[0,1].
\label{eq:code-similarity}
\end{equation}
Here, $\mathbf{I}$ is the identity matrix, and $\eta$ is the coefficient that controls the contribution of PPMI-based similarity.
Using Eq.~\eqref{eq:code-similarity}, we map a binary vector indicating which medical codes are present to a clinical feature that captures both these codes and their PPMI similarities to other medical codes.

\paragraph{Patient Target Distribution.}

For candidate patient $i$, let $\mathbf{d}^{\mathrm{pat}}_i\in\{0,1\}^{V}$ be a binary vector whose $k$-th entry indicates whether medical code $k$ appears in the original observed prefix.
We represent candidate patient $i$ by the clinical feature $\boldsymbol{\phi}^{\mathrm{pat}}_i=\mathbf{d}^{\mathrm{pat}}_i\mathbf{S}/\norm{\mathbf{d}^{\mathrm{pat}}_i\mathbf{S}}_2$ and construct the anchor feature $\boldsymbol{\phi}^{\mathrm{pat}}_b$ in the same way.
We use a Gaussian radial basis function (RBF) target to assign different weights with respect to clinical similarity.
For each anchor patient, we retain a fixed number of the most clinically similar candidates to prevent weak similarities from contributing to the target distribution.
Each retained candidate receives the raw target weight
\begin{equation}
w_{b,i}^{\mathrm{pat}}
=
\exp\!\left(
-\frac{\text{dist}(\boldsymbol{\phi}^{\mathrm{pat}}_b,\boldsymbol{\phi}^{\mathrm{pat}}_i)^2}
{2\left(\sigma^{\mathrm{pat}}\rho_b^{\mathrm{pat}}\right)^2}
\right).
\label{eq:patient-rbf-target}
\end{equation}
All remaining candidate patients receive zero target weight.
Here, $\text{dist}(\cdot,\cdot)$ denotes the Euclidean distance.
The local scale $\rho_b^{\mathrm{pat}}$ is estimated from a nearest-neighbor clinical neighborhood, and $\sigma^{\mathrm{pat}}$ controls the RBF bandwidth.

\paragraph{Visit Target Distribution.}

The patient objective models clinical similarity across patients, while the visit objective models changes over time within a single observed prefix.
We therefore restrict candidate visits to the same patient.
Because a visit contains few medical codes, weak similarities in $\mathbf{S}$ can make distinct visits appear similar.
We therefore retain a fixed number of the largest positive values in each row of $\mathbf{S}$ and denote the resulting sparse matrix by $\widehat{\mathbf{S}}$.
For visit $j$ of patient $b$, let $\mathbf{d}^{\mathrm{visit}}_{b,j}\in\{0,1\}^{V}$ be a binary vector whose $k$-th entry indicates whether medical code $k$ appears in the visit.
We represent the visit by the clinical feature $\boldsymbol{\phi}^{\mathrm{visit}}_{b,j}=\mathbf{d}^{\mathrm{visit}}_{b,j}\widehat{\mathbf{S}}/\norm{\mathbf{d}^{\mathrm{visit}}_{b,j}\widehat{\mathbf{S}}}_2$.

Visit intervals vary widely, from days to years.
We therefore apply a logarithmic transformation that preserves differences among short intervals while reducing the influence of long intervals.
For anchor visit $t$ and candidate visit $u$, the transformed time gap is
\begin{equation}
\Delta_{b,t,u}
=
\frac{\log\!\left(1+
\left|t_{b,t}^{\mathrm{day}}-t_{b,u}^{\mathrm{day}}\right|
\right)}{\kappa_v}.
\label{eq:visit-time-gap}
\end{equation}
We combine a clinical RBF weight with temporal decay.
Temporal decay assumes that clinically similar visits closer in time are more likely to belong to a related clinical episode.
Thus, clinically similar visits separated by shorter intervals receive larger target weights.
The resulting target weight for candidate visit $u$ is
{\medmuskip=2mu\thickmuskip=3mu
\begin{equation}
w_{b,t,u}^{\mathrm{visit}}
=
\exp\!\left(
-\frac{1}{2}
\left(
\frac{\text{dist}(\boldsymbol{\phi}^{\mathrm{visit}}_{b,t},\boldsymbol{\phi}^{\mathrm{visit}}_{b,u})}
{\sigma^{\mathrm{visit}}\rho_{b,t}^{\mathrm{visit}}}
\right)^2
-\frac{\Delta_{b,t,u}}{\gamma^{\mathrm{time}}}
\right).
\label{eq:visit-rbf-target}
\end{equation}
}

Here, $\text{dist}(\cdot,\cdot)$ denotes Euclidean distance, and $\rho_{b,t}^{\mathrm{visit}}$ is the local distance scale estimated from a nearest-neighbor clinical neighborhood.
The parameter $\sigma^{\mathrm{visit}}$ controls the RBF bandwidth.
The parameter $\kappa_v>0$ sets the time scale, and $\gamma^{\mathrm{time}}>0$ controls the rate of temporal decay.

\begin{table*}[t]
\centering
\caption{
Macro test AUROC across 14 EHRSHOT~\cite{ehrshot} tasks and label budgets, where finite $k$ denotes labeled prediction instances per class.
Values are mean $\pm$ SD over three pretraining seeds.
\textsuperscript{\dag} denotes an adaptation of an official implementation.
\textsuperscript{\ddag} denotes a re-implementation from the published method.
Best is in \textbf{bold}; second-best is \underline{underlined}.
}
\label{tab:ehrshot-main}
\tiny
\setlength{\tabcolsep}{1.0pt}
\begin{tabular*}{\textwidth}{@{\extracolsep{\fill}}lccccccc@{}}
\toprule
Method & \multicolumn{7}{c}{$k$} \\
\cmidrule(lr){2-8}
& 1 & 2 & 4 & 8 & 12 & 16 & 24 \\
\midrule
BEHRT (2020)\textsuperscript{\dag} \cite{behrt} & $0.524\pm0.002$ & $0.539\pm0.002$ & $0.565\pm0.004$ & $0.584\pm0.002$ & $0.590\pm0.002$ & $0.592\pm0.001$ & $0.596\pm0.001$ \\
CLMBR (2021)\textsuperscript{\ddag} \cite{clmbr} & $0.510\pm0.004$ & $0.501\pm0.006$ & $0.514\pm0.002$ & $0.510\pm0.002$ & $0.509\pm0.002$ & $0.513\pm0.003$ & $0.514\pm0.005$ \\
CEHR-BERT (2021)\textsuperscript{\dag} \cite{cehrbert} & $0.535\pm0.002$ & $0.546\pm0.003$ & $0.573\pm0.001$ & $0.595\pm0.001$ & $0.602\pm0.004$ & $0.605\pm0.003$ & $0.607\pm0.001$ \\
Med-BERT (2021)\textsuperscript{\dag} \cite{medbert} & $0.511\pm0.007$ & $0.510\pm0.007$ & $0.526\pm0.005$ & $0.529\pm0.004$ & $0.535\pm0.002$ & $0.538\pm0.003$ & $0.548\pm0.003$ \\
BYOS-EHR (2021)\textsuperscript{\ddag} \cite{wanyan2021positive} & $0.533\pm0.007$ & $0.550\pm0.003$ & $0.579\pm0.003$ & $0.596\pm0.001$ & $0.603\pm0.003$ & $0.612\pm0.002$ & $0.619\pm0.002$ \\
CLMBR-T (2023)\textsuperscript{\dag} \cite{ehrshot} & $0.547\pm0.002$ & $0.563\pm0.004$ & $0.585\pm0.003$ & $0.607\pm0.001$ & $0.614\pm0.002$ & $0.624\pm0.003$ & $0.629\pm0.002$ \\
TransformEHR (2023)\textsuperscript{\dag} \cite{transformehr} & $0.514\pm0.004$ & $0.509\pm0.002$ & $0.530\pm0.004$ & $0.538\pm0.004$ & $0.546\pm0.008$ & $0.554\pm0.005$ & $0.567\pm0.007$ \\
ETHOS (2024)\textsuperscript{\dag} \cite{ethos} & $0.548\pm0.003$ & $0.563\pm0.001$ & $0.591\pm0.004$ & $0.604\pm0.007$ & $0.608\pm0.004$ & $0.620\pm0.002$ & $0.631\pm0.002$ \\
MOTOR (2024)\textsuperscript{\dag} \cite{motor} & $0.512\pm0.001$ & $0.512\pm0.008$ & $0.549\pm0.002$ & $0.545\pm0.002$ & $0.561\pm0.002$ & $0.560\pm0.001$ & $0.564\pm0.001$ \\
EHRMamba (2025)\textsuperscript{\dag} \cite{ehrmamba} & $\mathbf{0.558\pm0.001}$ & $\mathbf{0.578\pm0.002}$ & $\underline{0.602\pm0.004}$ & $\underline{0.619\pm0.004}$ & $\underline{0.623\pm0.004}$ & $\underline{0.633\pm0.003}$ & $\underline{0.642\pm0.001}$ \\
TACT (2026)\textsuperscript{\ddag} \cite{tact} & $0.510\pm0.007$ & $0.510\pm0.005$ & $0.520\pm0.005$ & $0.528\pm0.005$ & $0.534\pm0.004$ & $0.539\pm0.006$ & $0.547\pm0.008$ \\
\midrule
EHR-CSS (ours) & $\underline{0.552\pm0.002}$ & $\underline{0.574\pm0.002}$ & $\mathbf{0.608\pm0.003}$ & $\mathbf{0.634\pm0.001}$ & $\mathbf{0.648\pm0.002}$ & $\mathbf{0.664\pm0.004}$ & $\mathbf{0.675\pm0.005}$ \\
$\Delta$ (ours $-$ second-best) & $-0.006$ & $-0.004$ & $+0.006$ & $+0.015$ & $+0.025$ & $+0.031$ & $+0.033$ \\
\bottomrule
\end{tabular*}
\par\medskip
\begin{tabular*}{\textwidth}{@{\extracolsep{\fill}}lcccccc@{}}
\toprule
Method & \multicolumn{6}{c}{$k$} \\
\cmidrule(lr){2-7}
& 32 & 48 & 64 & 128 & 256 & All \\
\midrule
BEHRT (2020)\textsuperscript{\dag} \cite{behrt} & $0.598\pm0.002$ & $0.604\pm0.001$ & $0.606\pm0.001$ & $0.616\pm0.005$ & $0.647\pm0.005$ & $0.679\pm0.007$ \\
CLMBR (2021)\textsuperscript{\ddag} \cite{clmbr} & $0.511\pm0.003$ & $0.515\pm0.001$ & $0.517\pm0.002$ & $0.534\pm0.001$ & $0.545\pm0.006$ & $0.555\pm0.007$ \\
CEHR-BERT (2021)\textsuperscript{\dag} \cite{cehrbert} & $0.608\pm0.001$ & $0.611\pm0.001$ & $0.614\pm0.004$ & $0.621\pm0.002$ & $0.641\pm0.004$ & $0.677\pm0.006$ \\
Med-BERT (2021)\textsuperscript{\dag} \cite{medbert} & $0.552\pm0.003$ & $0.555\pm0.005$ & $0.564\pm0.005$ & $0.581\pm0.004$ & $0.610\pm0.002$ & $0.638\pm0.003$ \\
BYOS-EHR (2021)\textsuperscript{\ddag} \cite{wanyan2021positive} & $0.622\pm0.004$ & $0.627\pm0.005$ & $0.632\pm0.007$ & $0.645\pm0.005$ & $0.667\pm0.003$ & $0.699\pm0.006$ \\
CLMBR-T (2023)\textsuperscript{\dag} \cite{ehrshot} & $0.632\pm0.001$ & $0.637\pm0.001$ & $0.638\pm0.003$ & $0.640\pm0.003$ & $\underline{0.687\pm0.003}$ & $\underline{0.719\pm0.006}$ \\
TransformEHR (2023)\textsuperscript{\dag} \cite{transformehr} & $0.573\pm0.010$ & $0.580\pm0.007$ & $0.582\pm0.008$ & $0.590\pm0.006$ & $0.601\pm0.005$ & $0.637\pm0.007$ \\
ETHOS (2024)\textsuperscript{\dag} \cite{ethos} & $0.637\pm0.003$ & $0.640\pm0.002$ & $0.646\pm0.003$ & $0.651\pm0.003$ & $0.675\pm0.004$ & $0.703\pm0.004$ \\
MOTOR (2024)\textsuperscript{\dag} \cite{motor} & $0.568\pm0.002$ & $0.564\pm0.001$ & $0.563\pm0.001$ & $0.559\pm0.001$ & $0.583\pm0.002$ & $0.602\pm0.009$ \\
EHRMamba (2025)\textsuperscript{\dag} \cite{ehrmamba} & $\underline{0.645\pm0.000}$ & $\underline{0.650\pm0.002}$ & $\underline{0.652\pm0.003}$ & $\underline{0.655\pm0.003}$ & $0.672\pm0.006$ & $0.707\pm0.003$ \\
TACT (2026)\textsuperscript{\ddag} \cite{tact} & $0.553\pm0.007$ & $0.557\pm0.006$ & $0.564\pm0.004$ & $0.583\pm0.008$ & $0.608\pm0.008$ & $0.644\pm0.007$ \\
\midrule
EHR-CSS (ours) & $\mathbf{0.685\pm0.005}$ & $\mathbf{0.699\pm0.003}$ & $\mathbf{0.707\pm0.004}$ & $\mathbf{0.722\pm0.002}$ & $\mathbf{0.728\pm0.003}$ & $\mathbf{0.744\pm0.003}$ \\
$\Delta$ (ours $-$ second-best) & $+0.040$ & $+0.049$ & $+0.055$ & $+0.067$ & $+0.041$ & $+0.025$ \\
\bottomrule
\end{tabular*}
\end{table*}

\subsubsection{Medical-Code Objectives}

Let $d\in\{1,\ldots,V\}$ index a candidate medical code, and let $\ell$ index a medical-code position in the serialized observed-prefix sequence.
For these objectives, $\bar{\mathbf{z}}_{b,\ell}^{\mathrm{code}}$ and $\bar{\mathbf{z}}_{b,\ell}^{\mathrm{ncode}}$ denote the corresponding visit-indexed representations from Eq.~\eqref{eq:objective-projections} after flattening $(t,\ell)$ into the sequence-position index $\ell$.
The three medical-code model distributions use the unit-normalized medical-code input embeddings $\bar{\mathbf{e}}_d=\mathbf{e}_d^{\mathrm{code}}/\norm{\mathbf{e}_d^{\mathrm{code}}}_2$ as candidate representations.
Their unnormalized model scores are
\begin{equation}
\begin{aligned}
\tilde q_{b, \ell, d}^{\mathrm{code}}
&=
\exp\!\left(
-\frac{1}{\tau_{\mathrm{code}}}
\norm{\bar{\mathbf{z}}_{b, \ell}^{\mathrm{code}}-\bar{\mathbf{e}}_d}^{2}
\right),
\\
\tilde q_{b, d}^{\mathrm{nvcode}}
&=
\exp\!\left(
-\frac{1}{\tau_{\mathrm{nvcode}}}
\norm{\bar{\mathbf{z}}_b^{\mathrm{nvcode}}-\bar{\mathbf{e}}_d}^{2}
\right),
\\
\tilde q_{b, \ell, d}^{\mathrm{ncode}}
&=
\exp\!\left(
-\frac{1}{\tau_{\mathrm{ncode}}}
\norm{\bar{\mathbf{z}}_{b, \ell}^{\mathrm{ncode}}-\bar{\mathbf{e}}_d}^{2}
\right).
\end{aligned}
\label{eq:code-unnormalized-model}
\end{equation}
Here, $\tau_{\mathrm{code}},\tau_{\mathrm{nvcode}},\tau_{\mathrm{ncode}}>0$ are temperatures.
The three medical-code objectives are defined as follows.

\paragraph{Masked-Code Target Distribution.}
The masked-code target distribution assigns positive weight only to the original medical code at each masked position.
For each masked position $\ell$, let $d_{b\ell}$ denote the original medical code.
The target weight assigned to candidate medical code $d$ is
\begin{equation}
w_{b, \ell, d}^{\mathrm{code}}
=
\mathbf 1\!\left\{d=d_{b, \ell}\right\}.
\label{eq:masked-code-target}
\end{equation}
Here, $\mathbf 1\!\left\{\cdot\right\}$ is the indicator function.

\paragraph{Next-Visit Code Target Distribution.}
The next-visit target distribution assigns positive weight to the medical codes in the withheld visit.
Common medical codes appear in many withheld visits and therefore contribute more often during pretraining.
We assign larger target weights to less frequent codes, while logarithmic scaling and smoothing prevent these weights from becoming excessively large.
Let $N$ be the number of pretraining patients, and let $n_d$ be the number of these patients with medical code $d$.
The smoothed inverse patient frequency is
\begin{equation}
\xi_d
=
\log\frac{N+1}{n_d+1}+1.
\label{eq:future-code-idf}
\end{equation}
Let $\mathcal D_b^{+}$ contain the distinct medical codes in the withheld visit.
The target weight for candidate medical code $d$ is
\begin{equation}
w_{b,d}^{\mathrm{nvcode}}
=
\mathbf 1\!\left\{d\in\mathcal D_b^{+}\right\}
\xi_d^{\beta},
\qquad \beta\geq0.
\label{eq:future-code-weight}
\end{equation}
Here, $\beta$ controls the influence of inverse patient frequency on the target weights.

\paragraph{Dense Next-Code Target Distribution.}
The dense next-code target distribution assigns positive weight only to the following medical code at each eligible position in the observed prefix.
We apply this target at every position $\ell$ followed by a valid medical code $d_{b,\ell+1}$.
The target weight for candidate medical code $d$ is
\begin{equation}
w_{b,\ell,d}^{\mathrm{ncode}}
=
\mathbf 1\!\left\{d=d_{b,\ell+1}\right\}.
\label{eq:dense-next-code-target}
\end{equation}
\subsection{Future Representation for Downstream Prediction}
\label{sec:downstream-representation}

For patient $b$ and prediction time $t^\star$, let $X_{b,t^\star}^{-}$ denote the patient history available before $t^\star$.
Let $c_b(t^\star)$ denote the index of the most recent visit in this history.
We encode $X_{b,t^\star}^{-}$ without masking and use Eq.~\eqref{eq:hierarchical-pooling} to obtain the patient representation $\mathbf{h}_{b,t^\star}^{\mathrm{pat}}$ and the most recent visit representation $\mathbf{h}_{b,c_b(t^\star)}^{\mathrm{visit}}$.
The next-visit projection produces the future representation
\begin{equation}
\begin{aligned}
\mathbf{z}_{b,t^\star}^{\mathrm{nvcode}}
&=
g^{\mathrm{nvcode}}\!\left(
[\mathbf{h}_{b,t^\star}^{\mathrm{pat}};
\mathbf{h}_{b,c_b(t^\star)}^{\mathrm{visit}}]
\right).
\end{aligned}
\label{eq:downstream-future-representation}
\end{equation}
Here, $[\cdot;\cdot]$ denotes concatenation.
The next-visit objective optimizes $\mathbf{z}_{b,t^\star}^{\mathrm{nvcode}}$ to predict the medical codes in the withheld visit.
We use this future representation as input to a task-specific linear classifier.

\section{Experiments}
\label{sec:experiments}

We compare EHR-CSS with established and state-of-the-art methods for longitudinal EHR representation learning \cite{behrt,cehrbert,medbert,clmbr,ehrshot,transformehr,ethos,motor,wanyan2021positive,tact,ehrmamba} on EHRSHOT \cite{ehrshot} and MIMIC-IV \cite{mimiciv}.
For both benchmarks, we report AUROC for each downstream task and its unweighted average across tasks.
Pretraining uses only the designated public data for each benchmark, with no external data provided to any model.
We use the same patients, data splits, medical codes, visit times, maximum sequence length, and number of pretraining epochs.
Published results based on unavailable datasets, external pretraining data, or incompatible task definitions are not directly comparable.
We therefore evaluate all methods with the same experimental setup.

We adapt official implementations when available and otherwise reimplement each method from its published specification within the shared structured-code pipeline.
Appendix~\ref{app:single-view-configuration} describes the datasets and benchmarks and provides details of data preprocessing, model implementation, pretraining, hyperparameter selection, and downstream evaluation.
The following subsections present results on EHRSHOT \cite{ehrshot} and MIMIC-IV \cite{mimiciv}.

\subsection{EHRSHOT Results}
\label{sec:ehrshot-results}

Table~\ref{tab:ehrshot-main} reports macro AUROC for $k\in\{1,2,4,8,12,16,24,32,48,64,128,256\}$ and under full supervision.
EHR-CSS achieves the highest AUROC at every label budget from $k=4$ through full supervision.
Its margin over the best baseline increases from $k=4$ through $k=128$ and remains positive at $k=256$ and under full supervision.
EHRMamba \cite{ehrmamba} performs best at $k=1$ and $k=2$, where EHR-CSS trails by $0.006$ and $0.004$.
At these budgets, the classifier is fitted on only two or four labeled instances in total.
We hypothesize that these instances capture too little within-class variation for the classifier to use the EHR-CSS features effectively.
Appendix~\ref{app:ehrshot-task-results} reports AUROC for each EHRSHOT task.
Under full supervision, EHR-CSS outperforms the best task-specific baseline on 10 of 14 tasks, with the largest gains on new-diagnosis tasks and the largest gaps on length of stay (LOS) and lupus.
We hypothesize that the patient, visit, and future-event objectives are particularly useful for new-diagnosis tasks, which may depend on clinical patterns accumulated across multiple visits.
Overall, the EHRSHOT \cite{ehrshot} results show that EHR-CSS improves downstream prediction across most label budgets and most tasks.

\subsection{MIMIC-IV Results}
\label{sec:mimic-results}

\begin{table}[t]
\centering
\caption{
AUROC for mortality, readmission, and length-of-stay (LOS) prediction on MIMIC-IV \cite{mimiciv} for EHR-CSS and selected representative baselines.
Every method uses the same pretraining setup.
Values are mean $\pm$ SD over three pretraining seeds.
Best is in \textbf{bold}; second-best is \underline{underlined}.
}
\label{tab:mimic4-main}
\scriptsize
\setlength{\tabcolsep}{2.0pt}
\resizebox{\columnwidth}{!}{%
\begin{tabular}{@{}lcccc@{}}
\toprule
Method & Mortality & Readmission & LOS $>7$ & Average \\
\midrule
BEHRT~\cite{behrt} & $0.763\pm0.002$ & $0.651\pm0.001$ & $0.648\pm0.000$ & $0.687\pm0.000$ \\
CLMBR-T~\cite{ehrshot} & $0.786\pm0.001$ & $\underline{0.680\pm0.003}$ & $0.668\pm0.003$ & $0.711\pm0.002$ \\
EHRMamba~\cite{ehrmamba} & $\underline{0.789\pm0.000}$ & $\mathbf{0.681\pm0.002}$ & $\underline{0.670\pm0.003}$ & $\underline{0.713\pm0.002}$ \\
\midrule
EHR-CSS (ours) & $\mathbf{0.790\pm0.002}$ & $0.679\pm0.002$ & $\mathbf{0.674\pm0.002}$ & $\mathbf{0.714\pm0.002}$ \\
\bottomrule
\end{tabular}%
}
\end{table}

Table \ref{tab:mimic4-main} compares EHR-CSS with selected baselines.
Appendix Table \ref{tab:mimic4-full-appendix} reports results for all evaluated baselines.
EHR-CSS achieves the highest average AUROC and the best performance for mortality and LOS in the withheld final admission.
EHRMamba \cite{ehrmamba} performs best for predicting whether the final admission occurs within 30 days of the preceding discharge.
As shown in Table \ref{tab:mimic4-main}, the improvement over the baselines is smaller than that observed on EHRSHOT \cite{ehrshot}.
MIMIC-IV \cite{mimiciv} contains hospital and ICU records, and the patient histories used here have fewer visits than those in EHRSHOT \cite{ehrshot}.
We hypothesize that these shorter histories reduce the benefit of the patient and visit objectives.
We also hypothesize that EHR-CSS performs less well on the readmission task because pretraining does not predict the time between admissions.
The readmission label is determined by the interval between discharge and the next admission.
In contrast, mortality and LOS describe the subsequent admission itself, and the next-visit objective is trained to predict medical codes from such future admissions.
Overall, EHR-CSS achieves the highest average MIMIC-IV \cite{mimiciv} result, but its improvement is smaller and varies across the three tasks.

\section{Ablation}
\label{sec:ablation}

\begin{table}[t]
\centering
\caption{
Objective, target-geometry, and divergence ablations.
EHRSHOT~\cite{ehrshot} results use $k=\mathrm{All}$, and MIMIC-IV~\cite{mimiciv} results are averaged across 3 tasks.
Leave-one-objective-out values are mean $\pm$ SD over three pretraining seeds.
Best is in \textbf{bold}.
}
\label{tab:objective-ablation}
{\footnotesize
\setlength{\tabcolsep}{3.5pt}
\renewcommand{\arraystretch}{1.08}
\resizebox{\columnwidth}{!}{%
\begin{tabular}{@{}lcccc@{}}
\toprule
& \multicolumn{2}{c}{EHRSHOT \cite{ehrshot}}
& \multicolumn{2}{c}{MIMIC-IV \cite{mimiciv}} \\
\cmidrule(lr){2-3}\cmidrule(l){4-5}
Configuration
& AUROC
& $\Delta$
& AUROC
& $\Delta$ \\
\midrule
\multicolumn{5}{@{}l}{\textit{Leave-One-Objective-Out}} \\
\addlinespace[2pt]
Full model & $\mathbf{0.744\pm0.003}$ & Ref. & $\mathbf{0.714\pm0.002}$ & Ref. \\
$-$ Patient & $0.725\pm0.007$ & $-0.019$ & $0.713\pm0.001$ & $-0.001$ \\
$-$ Visit & $0.732\pm0.007$ & $-0.012$ & $0.709\pm0.002$ & $-0.005$ \\
$-$ Masked code & $0.728\pm0.004$ & $-0.016$ & $0.707\pm0.001$ & $-0.007$ \\
$-$ Next-visit code & $0.729\pm0.005$ & $-0.015$ & $0.703\pm0.001$ & $-0.011$ \\
$-$ Dense next-code & $0.723\pm0.005$ & $-0.021$ & $0.702\pm0.002$ & $-0.012$ \\

\midrule
\multicolumn{5}{@{}l}{\textit{Target Geometry}} \\
\addlinespace[2pt]

Power target
& $0.730\pm0.001$
& Ref.
& $0.702\pm0.002$
& Ref. \\

RBF target
& $\mathbf{0.744\pm0.003}$
& $+0.014$
& $\mathbf{0.714\pm0.002}$
& $+0.012$ \\

\midrule
\multicolumn{5}{@{}l}{\textit{Divergences}} \\
\addlinespace[2pt]

KL
& $0.735\pm0.003$
& Ref.
& $0.697\pm0.002$
& Ref. \\

CSS
& $\mathbf{0.744\pm0.003}$
& $+0.009$
& $\mathbf{0.714\pm0.002}$
& $+0.017$ \\

\bottomrule
\end{tabular}%
}
}
\end{table}

Table~\ref{tab:objective-ablation} reports ablations of the pretraining objectives, target geometry, and distribution-alignment loss.
The leave-one-objective-out ablation removes one objective from the full model while retaining the other four.
Every removal lowers mean AUROC on both benchmarks, and EHRSHOT shows a larger decrease than MIMIC-IV for all five removals.
EHRSHOT contains more visits per patient than the MIMIC-IV cohort.
We hypothesize that more visits per patient help the patient, visit, and future-prediction objectives learn clinical patterns over time.
The complete model performs best on both benchmarks, supporting the use of all five objectives.

For the target geometry, we compare the locally scaled RBF targets in Eqs.~\eqref{eq:patient-rbf-target} and~\eqref{eq:visit-rbf-target} with power targets, which raise positive clinical-similarity scores to a fixed exponent before normalization.
Changing only the target construction from power targets to RBF targets increases AUROC.
The RBF targets adjust clinical distances according to the neighborhood of each anchor, whereas the power targets apply the same exponent to all positive similarity scores.
We hypothesize that this local adjustment assigns more suitable target probabilities to nearby and distant candidates, which improves prediction.

For distribution alignment, we replace the CSS loss with a forward KL-divergence loss while keeping the model distributions, target distributions, and remaining training setup unchanged.
For a fixed target distribution, forward KL differs from soft-label cross-entropy only by the target entropy, which has no gradient.
This comparison is therefore a same-target soft-label cross-entropy control.
CSS improves AUROC on both benchmarks, with a larger improvement on MIMIC-IV~\cite{mimiciv} than on EHRSHOT~\cite{ehrshot}.
This result supports using CSS to align model and target distributions across the patient, visit, and medical-code granularities.

\section{Discussion and Limitations}
\label{sec:discussion-limitations}

Our evaluation is limited by the scale and clinical coverage of available public EHR benchmarks.
EHRSHOT~\cite{ehrshot} covers diverse prediction tasks but contains a relatively small number of patients, while MIMIC-IV~\cite{mimiciv} is primarily an acute and critical care dataset.
Public EHR datasets that support both large-scale pretraining and standardized downstream evaluation remain limited.
Comparisons across published methods are further complicated by differences in their input and pretraining data.
Some methods use additional event types or numerical measurements \cite{motor,ethos}, while others are pretrained on larger institutional or external datasets \cite{behrt,ehrshot}.
To reduce these differences, we evaluate all methods using the same clinical input, patients, data splits, and number of pretraining epochs.
This setup supports a fair comparison but does not show how EHR-CSS performs with broader clinical information or substantially more pretraining data.
 Future work should evaluate EHR-CSS on larger multimodal datasets.

Our work focuses on the pretraining objectives rather than the sequence encoder architecture.
EHR-CSS uses a shared Transformer encoder for all five objectives, and we do not compare alternative encoder architectures.
More efficient encoders could process longer patient histories, which could provide more information to the patient objective and more visits to the visit objective.
 Future work should compare Transformer, recurrent, and state-space encoders under the same setup \cite{behrt,clmbr,ehrmamba}.

\section{Conclusion}

In this study, we introduced EHR-CSS, a self-supervised contrastive framework that learns longitudinal EHR representations by aligning each model distribution with a target distribution over candidate representations.
The framework applies the CSS objective at the patient, visit, and medical-code granularities to combine graded clinical relations with masked-code and future-event prediction.
EHR-CSS outperforms the evaluated state-of-the-art methods overall on the public datasets EHRSHOT \cite{ehrshot} and MIMIC-IV \cite{mimiciv}.

\section*{Acknowledgments}

This work was partially funded by the Research Council of Norway through Visual Intelligence grant no. 309439 and IKTPLUSS grant no. 303514. The work was also supported by the Pioneer Centre for AI under DNRF grant no. P1.

\section*{LLM Disclosure}

In this work, we used LLMs/AI tools only in an assistive role, to edit content produced by the authors.
Specifically, we used Claude Opus 5.5 (Anthropic) through Claude Code to polish the language and structure of our own writing.
We also used it to assist with writing code that follows the method and experimental design described in this paper and the cited prior work.
We have not used LLMs/AI tools to generate synthetic data, theoretical or conceptual frameworks, mathematical claims or proofs, hypotheses, methodology or experimental design, data cleaning, interpretation of results, related work, citations, or figures.
Translation and qualitative data analysis are not applicable to this work.
We have reviewed all AI-assisted work.
We compared all edited text with our original draft and reviewed and tested all code before use.
We take responsibility for the final content of this work, including text, claims, or artifacts produced with the aid of LLMs/AI tools.

\printbibliography

\clearpage
\appendix

\section{Derivation of the CSS Alignment and Uniformity Terms}
\label{app:css-decomposition}

For a fixed anchor $i$, the discrete Cauchy--Schwarz divergence \cite{principe2000itl,jenssen2006cs} between the target distribution $\mathbf{p}_i$ and model distribution $\mathbf{q}_i$ is
\begin{equation}
D_{\mathrm{CS}}(\mathbf{p}_i,\mathbf{q}_i)
=
-\log
\frac{
\left(\sum_{j=1}^{T}p_{ij}q_{ij}\right)^2
}{
\left(\sum_{j=1}^{T}p_{ij}^{2}\right)
\left(\sum_{j=1}^{T}q_{ij}^{2}\right)
}.
\label{eq:appendix-cs-divergence}
\end{equation}
EHR-CSS uses one half of this divergence as its row loss.
This scaling preserves the minimizer and gives
\begin{equation}
\ell_{\mathrm{CSS}}(\mathbf{p}_i,\mathbf{q}_i)
=
-\log
\frac{
\sum_{j=1}^{T}p_{ij}q_{ij}
}{
\left(\sum_{j=1}^{T}p_{ij}^{2}\right)^{1/2}
\left(\sum_{j=1}^{T}q_{ij}^{2}\right)^{1/2}
}.
\label{eq:appendix-cs-start}
\end{equation}
The model distribution satisfies $q_{ij}=\tilde q_{ij}/Z_i$, where $Z_i>0$.
The numerator and the model norm can therefore be written as
\begin{align}
\sum_{j=1}^{T}p_{ij}q_{ij}
&=
\frac{1}{Z_i}
\sum_{j=1}^{T}p_{ij}\tilde q_{ij},
\label{eq:appendix-cs-numerator}
\\
\left(\sum_{j=1}^{T}q_{ij}^{2}\right)^{1/2}
&=
\frac{1}{Z_i}
\left(\sum_{j=1}^{T}\tilde q_{ij}^{2}\right)^{1/2}.
\label{eq:appendix-cs-model-norm}
\end{align}
Substituting Eqs. \eqref{eq:appendix-cs-numerator} and \eqref{eq:appendix-cs-model-norm} into Eq. \eqref{eq:appendix-cs-start} cancels the common factor $Z_i^{-1}$ and gives
\begin{equation}
\ell_{\mathrm{CSS}}(\mathbf{p}_i,\mathbf{q}_i)
=
-\log
\frac{
\sum_{j=1}^{T}p_{ij}\tilde q_{ij}
}{
\left(\sum_{j=1}^{T}p_{ij}^{2}\right)^{1/2}
\left(\sum_{j=1}^{T}\tilde q_{ij}^{2}\right)^{1/2}
}.
\label{eq:appendix-cs-cancelled}
\end{equation}
Applying $-\log(a/(bc))=-\log a+\log b+\log c$ separates the numerator from the two denominator factors.
The square-root powers are then moved outside the logarithm
\begin{equation}
\begin{aligned}
\ell_{\mathrm{CSS}}(\mathbf{p}_i,\mathbf{q}_i)
&=
-\log\!\left(\sum_{j=1}^{T}p_{ij}\tilde q_{ij}\right)
\\
&\quad+
\log\!\left(\sum_{j=1}^{T}\tilde q_{ij}^{2}\right)^{1/2}
\\
&\quad+
\log\!\left(\sum_{j=1}^{T}p_{ij}^{2}\right)^{1/2}
\\[2pt]
&=
-\log\sum_{j=1}^{T}p_{ij}\tilde q_{ij}
\\
&\quad+
\frac{1}{2}\log\sum_{j=1}^{T}\tilde q_{ij}^{2}
\\
&\quad+
\frac{1}{2}\log\sum_{j=1}^{T}p_{ij}^{2}.
\end{aligned}
\label{eq:appendix-cs-expanded}
\end{equation}
The target distribution $\mathbf{p}_i$ is fixed with respect to the model parameters.
Its contribution
\begin{equation}
C_i
=
\frac{1}{2}\log\sum_{j=1}^{T}p_{ij}^{2}
\label{eq:appendix-target-constant}
\end{equation}
is therefore constant during optimization.
The exact decomposition is
\begin{equation}
\begin{aligned}
\ell_{\mathrm{CSS}}(\mathbf{p}_i,\mathbf{q}_i)
&=
\underbrace{
-\log\sum_{j=1}^{T}p_{ij}\tilde q_{ij}
}_{\mathcal L_{i,\mathrm{align}}}
\\
&\quad+
\underbrace{
\frac{1}{2}\log\sum_{j=1}^{T}\tilde q_{ij}^{2}
}_{\mathcal L_{i,\mathrm{uniform}}}
+C_i.
\end{aligned}
\label{eq:appendix-cs-alignment-uniformity}
\end{equation}
Using the distance-based model weight from Eq. \eqref{eq:unnormalized-model}, the two parameter-dependent terms are
\begin{align}
\mathcal L_{i,\mathrm{align}}
&=
-\log\sum_{j=1}^{T}p_{ij}
\exp\!\left(
-\frac{1}{\tau}\norm{\bar{\mathbf{z}}_i-\bar{\mathbf{z}}_j}^{2}
\right),
\label{eq:appendix-alignment-distance}
\\
\mathcal L_{i,\mathrm{uniform}}
&=
\frac{1}{2}\log\sum_{j=1}^{T}
\exp\!\left(
-\frac{2}{\tau}\norm{\bar{\mathbf{z}}_i-\bar{\mathbf{z}}_j}^{2}
\right).
\label{eq:appendix-uniformity-distance}
\end{align}

\section{PPMI Medical-Code Association Matrix}
\label{app:ppmi}

Let $N$ denote the number of patients.
We estimate the PPMI \cite{church1990word, bullinaria2007extracting} matrix $\mathbf{A}\in \mathbb{R}^{|V|\times |V|}$ from patient-level medical-code co-occurrence in the pretraining records. The PPMI between medical-code $d\in V$ and $e\in V$ is estimated via 
\begin{equation}
A_{d,e}
=
\begin{cases}
\displaystyle
\left[
\log
\frac{n_{d\mathbin{\&}e}/N+\epsilon}
{(n_d/N)(n_e/N)+\epsilon}
\right]_{+},
& d\neq e,
\\[8pt]
0,
& d=e,
\end{cases}
\label{eq:appendix-ppmi}
\end{equation}
where $n_d$ and $n_e$ are the numbers of patients containing medical codes $d$ and $e$, respectively.
The term $n_{d\mathbin{\&}e}$ is the number of patients containing both medical codes, and $\epsilon>0$ is a small numerical constant.
The operator $[x]_{+}=\max(x,0)$ retains only positive values.
The diagonal is set to zero because exact medical-code identity is included separately in Eq.~\eqref{eq:code-similarity}.
We normalize each nonzero row of $\mathbf{A}$ to sum to one as
\begin{equation}
\widetilde A_{d,e}
=
\begin{cases}
\displaystyle
\frac{A_{d,e}}{\sum_{k=1}^{V}A_{d,k}},
& \displaystyle \sum_{k=1}^{V}A_{d,k}>0,
\\[8pt]
0,
& \text{otherwise}.
\end{cases}
\label{eq:appendix-ppmi-row-normalization}
\end{equation}

\section{Implementation and Evaluation Details}
\label{app:single-view-configuration}

\subsection{Datasets and Benchmarks}
\label{app:datasets-benchmarks}

\paragraph{EHRSHOT~\cite{ehrshot}.}
EHRSHOT is a few-shot benchmark based on patient records from the Stanford Medicine Research Data Repository represented using the OMOP common data model.
It includes 6,739 patients and fixed training, validation, and test sets.
We evaluate 14 tasks defined over structured diagnosis, procedure, drug, and laboratory records.
These comprise 3 operational outcomes, 5 laboratory outcomes, and 6 new-diagnosis outcomes.
We exclude Chest X-Ray Findings, the benchmark's only multilabel task, because the shared downstream evaluation assumes one label per prediction event.

\paragraph{MIMIC-IV~\cite{mimiciv}.}
MIMIC-IV is a deidentified critical care database containing longitudinal hospital records.
We map diagnosis codes from each admission to Clinical Classifications Software categories~\cite{ahrqccs}.
We evaluate in-hospital mortality, length of stay (LOS) longer than 7 days, and readmission within 30 days.
The final admission defines mortality and LOS.
Readmission is defined by the interval between the preceding discharge and final admission.

\subsection{Data Preprocessing}
\label{app:data-construction}

For EHRSHOT~\cite{ehrshot}, pretraining and vocabulary construction use only the released training and validation sets.
Events at or after the prediction time are excluded from each downstream input.
For MIMIC-IV~\cite{mimiciv}, only the training set is used for pretraining, vocabulary construction, and model selection.
For downstream model selection, $20\%$ of this training set is used for validation.
The final admission is excluded from each downstream input, leaving only the patient's earlier history.
Table~\ref{tab:data-construction} lists the dataset-specific cutoff ranges and input limits.
At each pretraining epoch, we resample the cutoff $c_b$ and withhold the following visit $v_b^{+}$ as the next-visit target.
Each pretraining example contains one observed prefix without medical-code dropout.
For the masked-code objective, selected medical codes are replaced with a masking token.
The limits on visits and medical-code tokens control memory use, while the visit limit also bounds within-patient visit comparisons.
If an observed prefix exceeds either limit, only the most recent visits and medical-code tokens are retained.

\begin{table}[t]
\centering
\caption{Data preprocessing parameters for EHRSHOT~\cite{ehrshot} and MIMIC-IV~\cite{mimiciv}.}
\label{tab:data-construction}
\small
\setlength{\tabcolsep}{4pt}
\renewcommand{\arraystretch}{1.03}
\begin{tabular}{@{}lcc@{}}
\toprule
Setup & EHRSHOT & MIMIC-IV \\
\midrule
Minimum visits & $5$ & $5$ \\
$c_b$ range & $[0.50m_b,0.90m_b]$ & $[0.30m_b,0.70m_b]$ \\
Maximum visits & $128$ & $64$ \\
Maximum tokens & $512$ & $100$ \\
\bottomrule
\end{tabular}
\end{table}

\subsection{Model and Pretraining Implementation}
\label{app:implementation-details}

Following BEHRT~\cite{behrt}, EHR-CSS uses a 6-layer Transformer with hidden width $288$, $12$ attention heads, and feed-forward width $512$.
Hidden-state and attention dropout are both set to $0.10$.
Table~\ref{tab:pretraining-implementation} lists the dataset-specific projection dimensions, target-distribution parameters, objective weights, and optimization setup.

\begin{table}[t]
\centering
\caption{EHR-CSS pretraining setup for EHRSHOT~\cite{ehrshot} and MIMIC-IV~\cite{mimiciv}.}
\label{tab:pretraining-implementation}
\small
\setlength{\tabcolsep}{4pt}
\renewcommand{\arraystretch}{1.03}
\begin{tabular}{@{}lcc@{}}
\toprule
Setup & EHRSHOT & MIMIC-IV \\
\midrule
\shortstack[l]{Patient and visit\\projection width} & $256$ & $64$ \\
Patient queue size & $4096$ & $2048$ \\
$\eta$ & $0.50$ & $0.75$ \\
Patient candidate top-$k$ & $24$ & $32$ \\
Patient local-scale rank & $8$ & $8$ \\
$\sigma^{\mathrm{pat}}$ & $0.50$ & $1.00$ \\
Visit candidate top-$k$ & $32$ & All \\
Visit local-scale rank & $8$ & $4$ \\
$\sigma^{\mathrm{visit}}$ & $1.50$ & $1.00$ \\
Visit similarity threshold & $0.05$ & $0.02$ \\
$\kappa_v$ & $8$ & $8$ \\
$\gamma_{\mathrm{time}}$ & $0.25$ & $0.50$ \\
$\tau$ & $0.20$ & $0.10$ \\
Masking probability & $0.15$ & $0.10$ \\
Maximum masked codes & $48$ & $96$ \\
$\tau_{\mathrm{code}}$ & $0.10$ & $0.40$ \\
$\tau_{\mathrm{nvcode}}$ & $0.10$ & $0.20$ \\
$\tau_{\mathrm{ncode}}$ & $0.10$ & $0.40$ \\
$\beta_{\mathrm{nvcode}}$ & $1.00$ & $0$ \\
$\lambda_{\mathrm{pat}}$ & $0.75$ & $0.50$ \\
$\lambda_{\mathrm{vis}}$ & $0.50$ & $1.00$ \\
$\lambda_{\mathrm{code}}$ & $2.00$ & $4.00$ \\
$\lambda_{\mathrm{nvcode}}$ & $0.25$ & $0.10$ \\
$\lambda_{\mathrm{ncode}}$ & $0.50$ & $0.50$ \\
Pretraining epochs & $150$ & $150$ \\
Per-step batch size & $48$ & $64$ \\
Gradient accumulation & $2$ & $1$ \\
Learning rate & $10^{-4}$ & $2\mathord{\times}10^{-4}$ \\
Warm-up ratio & $0.20$ & $0.10$ \\
Maximum gradient norm & $1.00$ & $0.50$ \\
\bottomrule
\end{tabular}
\end{table}

We optimize EHR-CSS with BertAdam~\cite{behrt} and apply weight decay of $0.05$ to all parameters except bias and LayerNorm parameters.

\subsection{Hyperparameter Selection}
\label{app:hyperparameter-selection}

We select all hyperparameters using validation data.
For each benchmark, we conduct a pretraining hyperparameter search over the optimization setup, model dimensions, objective weights, masking, and target-distribution parameters.
Configurations are ranked by downstream validation AUROC, and MIMIC-IV~\cite{mimiciv} uses a separate validation search for fine-tuning hyperparameters.
For the EHRSHOT~\cite{ehrshot} target-distribution parameters, we use a scrambled Sobol search over the values in Table~\ref{tab:target-search}.
The search evaluates 24 configurations: 22 Sobol samples, one RBF reference, and one power-target control.
Each configuration is pretrained for 50 epochs with one seed and ranked by validation AUROC at $k\in\{16,32,64\}$.
All other parameters remain fixed during this search.
The selected configuration is then pretrained for 150 epochs.
All baselines use the same data, splits, input, and number of pretraining epochs.
We do not claim that the baselines received comparable tuning effort.
After selection, we repeat pretraining with three random initializations for each benchmark.

\begin{table}[t]
\centering
\caption{EHRSHOT~\cite{ehrshot} target-distribution search space. Selected values are listed in Table~\ref{tab:pretraining-implementation}.}
\label{tab:target-search}
\small
\setlength{\tabcolsep}{4pt}
\begin{tabular}{@{}ll@{}}
\toprule
Parameter & Searched values \\
\midrule
$\lambda_{\mathrm{pat}}$ & $\{0.15, 0.25, 0.50, 0.75, 1.00\}$ \\
Patient candidate top-$k$ & $\{4, 8, 16, 24, 32\}$ \\
Patient local-scale rank & $\{2, 4, 8, 16\}$ \\
$\sigma^{\mathrm{pat}}$ & $\{0.25, 0.50, 0.75, 1.00, 1.50\}$ \\
$\lambda_{\mathrm{vis}}$ & $\{0.25, 0.50, 0.75, 1.00, 1.50\}$ \\
Visit candidate top-$k$ & $\{\mathrm{All}, 4, 8, 16, 32\}$ \\
Visit local-scale rank & $\{2, 4, 8, 16\}$ \\
$\sigma^{\mathrm{visit}}$ & $\{0.25, 0.50, 0.75, 1.00, 1.50\}$ \\
Visit similarity threshold & $\{0, 0.02, 0.05, 0.10\}$ \\
$\gamma_{\mathrm{time}}$ & $\{0.25, 0.50, 1.00, 2.00\}$ \\
$\tau$ & $\{0.05, 0.075, 0.10, 0.15, 0.20\}$ \\
\bottomrule
\end{tabular}
\end{table}

\subsection{Downstream Evaluation}
\label{app:downstream-evaluation}

For EHRSHOT~\cite{ehrshot}, we follow the original CLMBR-T evaluation and fit logistic regression on frozen representations at every label budget, including All.
We freeze the pretrained sequence encoder and next-visit projection and use the $288$-dimensional future representation defined in Section~\ref{sec:downstream-representation}.
For each task and label budget, we fit a separate logistic regression classifier with $\ell_2$ regularization, $C=1$, and at most $2000$ iterations.
Evaluation covers label budgets $k\in\{1,2,4,8,12,16,24,32,48,64,128,256\}$ and full supervision.
All methods use the same sampled training examples.

For MIMIC-IV~\cite{mimiciv}, we initialize the sequence encoder and next-visit projection from the pretrained model, add a linear task head to the future representation, and fine-tune all components jointly.
For each task, we retrain the selected setup on the full training set and evaluate it once on the test set.
Label efficiency is therefore evaluated only on EHRSHOT~\cite{ehrshot}.

\section{Additional Results}
\label{app:additional-results}

\subsection{Task-Level EHRSHOT Results}
\label{app:ehrshot-task-results}

Table~\ref{tab:ehrshot-task-level} expands the EHR-CSS macro averages in Table~\ref{tab:ehrshot-main} into the 14 constituent EHRSHOT \cite{ehrshot} outcomes.
The task-level evaluation follows the same label budgets and frozen logistic-regression setup as the main comparison.

\begin{table*}[p]
\centering
\caption{
Task-level EHR-CSS test AUROC across 14 EHRSHOT~\cite{ehrshot} tasks and label budgets.
Values are mean $\pm$ SD over three pretraining seeds.
}
\label{tab:ehrshot-task-level}
\tiny
\setlength{\tabcolsep}{1.0pt}
\begin{tabular*}{\textwidth}{@{\extracolsep{\fill}}lccccccc@{}}
\toprule
Task & \multicolumn{7}{c}{Downstream label budget} \\
\cmidrule(lr){2-8}
& 1 & 2 & 4 & 8 & 12 & 16 & 24 \\
\midrule
\multicolumn{8}{l}{\textit{Operational outcomes}} \\
ICU admission & $0.682\pm0.004$ & $0.755\pm0.014$ & $0.819\pm0.011$ & $0.868\pm0.007$ & $0.892\pm0.007$ & $0.904\pm0.005$ & $0.913\pm0.007$ \\
Length of stay & $0.521\pm0.012$ & $0.599\pm0.008$ & $0.615\pm0.005$ & $0.662\pm0.003$ & $0.716\pm0.005$ & $0.738\pm0.009$ & $0.765\pm0.009$ \\
Readmission & $0.566\pm0.019$ & $0.584\pm0.013$ & $0.676\pm0.015$ & $0.696\pm0.004$ & $0.703\pm0.004$ & $0.727\pm0.005$ & $0.747\pm0.006$ \\
\addlinespace[2pt]
\multicolumn{8}{l}{\textit{Anticipating Lab Test Results}} \\
Anemia & $0.590\pm0.002$ & $0.609\pm0.009$ & $0.636\pm0.008$ & $0.668\pm0.008$ & $0.698\pm0.004$ & $0.717\pm0.005$ & $0.720\pm0.007$ \\
Hyperkalemia & $0.509\pm0.002$ & $0.529\pm0.004$ & $0.548\pm0.005$ & $0.588\pm0.001$ & $0.595\pm0.005$ & $0.587\pm0.007$ & $0.616\pm0.003$ \\
Hypoglycemia & $0.521\pm0.008$ & $0.533\pm0.010$ & $0.535\pm0.003$ & $0.513\pm0.003$ & $0.521\pm0.001$ & $0.533\pm0.002$ & $0.548\pm0.004$ \\
Hyponatremia & $0.497\pm0.002$ & $0.489\pm0.002$ & $0.520\pm0.006$ & $0.531\pm0.003$ & $0.544\pm0.004$ & $0.543\pm0.005$ & $0.532\pm0.004$ \\
Thrombocytopenia & $0.536\pm0.010$ & $0.540\pm0.013$ & $0.605\pm0.025$ & $0.638\pm0.024$ & $0.656\pm0.025$ & $0.673\pm0.023$ & $0.684\pm0.020$ \\
\addlinespace[2pt]
\multicolumn{8}{l}{\textit{Assignment of New Diagnoses}} \\
Acute myocardial infarction & $0.545\pm0.008$ & $0.513\pm0.008$ & $0.601\pm0.011$ & $0.583\pm0.016$ & $0.597\pm0.009$ & $0.618\pm0.012$ & $0.653\pm0.010$ \\
Celiac disease & $0.485\pm0.009$ & $0.519\pm0.016$ & $0.557\pm0.013$ & $0.581\pm0.009$ & $0.601\pm0.017$ & $0.622\pm0.027$ & $0.638\pm0.034$ \\
Hyperlipidemia & $0.515\pm0.011$ & $0.535\pm0.018$ & $0.555\pm0.003$ & $0.601\pm0.008$ & $0.602\pm0.004$ & $0.619\pm0.005$ & $0.624\pm0.005$ \\
Hypertension & $0.567\pm0.017$ & $0.600\pm0.013$ & $0.594\pm0.007$ & $0.622\pm0.008$ & $0.634\pm0.008$ & $0.644\pm0.009$ & $0.649\pm0.006$ \\
Lupus & $0.494\pm0.017$ & $0.510\pm0.014$ & $0.493\pm0.010$ & $0.547\pm0.025$ & $0.544\pm0.040$ & $0.580\pm0.037$ & $0.571\pm0.041$ \\
Pancreatic cancer & $0.699\pm0.012$ & $0.721\pm0.009$ & $0.760\pm0.014$ & $0.779\pm0.012$ & $0.771\pm0.016$ & $0.785\pm0.021$ & $0.786\pm0.022$ \\
\midrule
\textbf{Macro average} & $\mathbf{0.552\pm0.002}$ & $\mathbf{0.574\pm0.002}$ & $\mathbf{0.608\pm0.003}$ & $\mathbf{0.634\pm0.001}$ & $\mathbf{0.648\pm0.002}$ & $\mathbf{0.664\pm0.004}$ & $\mathbf{0.675\pm0.005}$ \\
\bottomrule
\end{tabular*}
\par\medskip
\begin{tabular*}{\textwidth}{@{\extracolsep{\fill}}lcccccc@{}}
\toprule
Task & \multicolumn{6}{c}{Downstream label budget} \\
\cmidrule(lr){2-7}
& 32 & 48 & 64 & 128 & 256 & All \\
\midrule
\multicolumn{7}{l}{\textit{Operational outcomes}} \\
ICU admission & $0.917\pm0.007$ & $0.921\pm0.007$ & $0.925\pm0.009$ & $0.932\pm0.011$ & $0.934\pm0.011$ & $0.937\pm0.010$ \\
Length of stay & $0.779\pm0.011$ & $0.804\pm0.006$ & $0.819\pm0.006$ & $0.832\pm0.005$ & $0.841\pm0.005$ & $0.862\pm0.004$ \\
Readmission & $0.756\pm0.005$ & $0.760\pm0.005$ & $0.771\pm0.003$ & $0.774\pm0.003$ & $0.777\pm0.004$ & $0.782\pm0.007$ \\
\addlinespace[2pt]
\multicolumn{7}{l}{\textit{Anticipating Lab Test Results}} \\
Anemia & $0.729\pm0.005$ & $0.751\pm0.004$ & $0.762\pm0.006$ & $0.785\pm0.005$ & $0.790\pm0.004$ & $0.817\pm0.003$ \\
Hyperkalemia & $0.638\pm0.003$ & $0.658\pm0.004$ & $0.657\pm0.001$ & $0.679\pm0.004$ & $0.696\pm0.003$ & $0.709\pm0.008$ \\
Hypoglycemia & $0.555\pm0.002$ & $0.563\pm0.002$ & $0.569\pm0.004$ & $0.588\pm0.007$ & $0.598\pm0.005$ & $0.621\pm0.008$ \\
Hyponatremia & $0.534\pm0.003$ & $0.548\pm0.004$ & $0.552\pm0.005$ & $0.578\pm0.007$ & $0.592\pm0.004$ & $0.628\pm0.006$ \\
Thrombocytopenia & $0.696\pm0.014$ & $0.726\pm0.008$ & $0.731\pm0.008$ & $0.747\pm0.002$ & $0.748\pm0.005$ & $0.773\pm0.006$ \\
\addlinespace[2pt]
\multicolumn{7}{l}{\textit{Assignment of New Diagnoses}} \\
Acute myocardial infarction & $0.674\pm0.012$ & $0.675\pm0.011$ & $0.687\pm0.011$ & $0.713\pm0.012$ & $0.717\pm0.005$ & $0.725\pm0.002$ \\
Celiac disease & $0.685\pm0.032$ & $0.705\pm0.022$ & $0.708\pm0.020$ & $0.709\pm0.020$ & $0.724\pm0.015$ & $0.752\pm0.023$ \\
Hyperlipidemia & $0.632\pm0.010$ & $0.642\pm0.010$ & $0.645\pm0.007$ & $0.659\pm0.006$ & $0.665\pm0.004$ & $0.672\pm0.005$ \\
Hypertension & $0.639\pm0.005$ & $0.658\pm0.007$ & $0.668\pm0.007$ & $0.674\pm0.008$ & $0.657\pm0.003$ & $0.667\pm0.007$ \\
Lupus & $0.559\pm0.032$ & $0.560\pm0.030$ & $0.578\pm0.030$ & $0.607\pm0.029$ & $0.626\pm0.050$ & $0.645\pm0.041$ \\
Pancreatic cancer & $0.804\pm0.018$ & $0.817\pm0.015$ & $0.826\pm0.015$ & $0.827\pm0.011$ & $0.829\pm0.013$ & $0.832\pm0.021$ \\
\midrule
\textbf{Macro average} & $\mathbf{0.685\pm0.005}$ & $\mathbf{0.699\pm0.003}$ & $\mathbf{0.707\pm0.004}$ & $\mathbf{0.722\pm0.002}$ & $\mathbf{0.728\pm0.003}$ & $\mathbf{0.744\pm0.003}$ \\
\bottomrule
\end{tabular*}
\end{table*}

\subsection{Complete MIMIC-IV Results}
\label{app:mimic4-results}

For MIMIC-IV \cite{mimiciv}, Table~\ref{tab:mimic4-full-appendix} extends Table~\ref{tab:mimic4-main} with every baseline evaluated under the same pretraining and evaluation setup.

\begin{table*}[t]
\centering
\caption{
Complete MIMIC-IV~\cite{mimiciv} test AUROC comparison.
Values are mean $\pm$ SD over three pretraining seeds.
\textsuperscript{\dag} and \textsuperscript{\ddag} denote adapted official implementations and reimplementations, respectively.
Best and second-best results are \textbf{bold} and \underline{underlined}.
}
\label{tab:mimic4-full-appendix}
\scriptsize
\begin{tabular*}{\textwidth}{@{\extracolsep{\fill}}lcccc@{}}
\toprule
Method & Mortality & Readmission & LOS $>7$ & Average \\
\midrule
BEHRT (2020)\textsuperscript{\(\dagger\)} \cite{behrt} & $0.763\pm0.002$ & $0.651\pm0.001$ & $0.648\pm0.000$ & $0.687\pm0.000$ \\
CLMBR (2021)\textsuperscript{\(\ddagger\)} \cite{clmbr} & $0.727\pm0.003$ & $0.624\pm0.002$ & $0.636\pm0.001$ & $0.662\pm0.000$ \\
CEHR-BERT (2021)\textsuperscript{\(\dagger\)} \cite{cehrbert} & $0.763\pm0.005$ & $0.656\pm0.004$ & $0.651\pm0.003$ & $0.690\pm0.004$ \\
Med-BERT (2021)\textsuperscript{\(\dagger\)} \cite{medbert} & $0.760\pm0.001$ & $0.655\pm0.002$ & $0.657\pm0.001$ & $0.691\pm0.001$ \\
BYOS-EHR (2021)\textsuperscript{\(\ddagger\)} \cite{wanyan2021positive} & $0.761\pm0.004$ & $0.654\pm0.000$ & $0.647\pm0.002$ & $0.687\pm0.002$ \\
CLMBR-T (2023)\textsuperscript{\(\dagger\)} \cite{ehrshot} & $0.786\pm0.001$ & $\underline{0.680\pm0.003}$ & $0.668\pm0.003$ & $0.711\pm0.002$ \\
TransformEHR (2023)\textsuperscript{\(\dagger\)} \cite{transformehr} & $0.778\pm0.003$ & $0.667\pm0.001$ & $0.665\pm0.001$ & $0.703\pm0.000$ \\
ETHOS (2024)\textsuperscript{\(\dagger\)} \cite{ethos} & $0.780\pm0.001$ & $0.671\pm0.002$ & $\underline{0.670\pm0.003}$ & $0.707\pm0.001$ \\
MOTOR (2024)\textsuperscript{\(\dagger\)} \cite{motor} & $0.774\pm0.003$ & $0.679\pm0.002$ & $0.649\pm0.002$ & $0.701\pm0.002$ \\
EHRMamba (2025)\textsuperscript{\(\dagger\)} \cite{ehrmamba} & $\underline{0.789\pm0.000}$ & $\mathbf{0.681\pm0.002}$ & $\underline{0.670\pm0.003}$ & $\underline{0.713\pm0.002}$ \\
TACT (2026)\textsuperscript{\(\ddagger\)} \cite{tact} & $0.752\pm0.003$ & $0.660\pm0.001$ & $0.644\pm0.002$ & $0.686\pm0.001$ \\
\midrule
EHR-CSS (ours)
& $\mathbf{0.790\pm0.002}$
& $0.679\pm0.002$
& $\mathbf{0.674\pm0.002}$
& $\mathbf{0.714\pm0.002}$ \\
\bottomrule
\end{tabular*}
\end{table*}

\end{document}